\documentclass[10pt,twocolumn,letterpaper]{article}

\usepackage{cvpr}
\usepackage{fontspec}
\usepackage{algorithm}
\usepackage{algpseudocode}
\usepackage{xcolor}
\usepackage[most]{tcolorbox}
\usepackage{microtype}
\usepackage{hyperref}
\usepackage{tikz}
\usetikzlibrary{fit,positioning,arrows.meta}

\definecolor{algFitTitle}{RGB}{230, 242, 235}
\definecolor{algFitBody}{RGB}{244, 250, 246}
\definecolor{algReassignTitle}{RGB}{235, 235, 250}
\definecolor{algReassignBody}{RGB}{246, 246, 252}
\definecolor{algFn}{RGB}{0, 110, 120}
\newcommand{\algfn}[1]{\textcolor{algFn}{\texttt{#1}}}
\tcbset{
  algblock/.style={
    boxrule=0pt,
    sharp corners,
    left=1.5mm,
    right=1.5mm,
    top=1mm,
    bottom=1mm,
    toptitle=1.5mm,
    bottomtitle=1.5mm,
    before skip=1.5mm,
    after skip=1.5mm,
    fontupper=\small,
    fonttitle=\small,
    coltitle=black,
  },
}

\title{\textbf{ImprovedVBGS}: Real-time Continual Variational\\
Bayes Gaussian Splatting}

\author{%
  Damani Mguni-Coker\\
  \texttt{Independent Researcher}\thanks{Project code:
  \href{https://github.com/damanimc/ImprovedVBGS}{\texttt{github.com/damanimc/ImprovedVBGS}}.}%
}

\date{}
\begin{document}
\maketitle
\begin{abstract}
On-the-fly reconstruction is a key requirement for many applications in robotics and autonomous navigation. Variational Bayes Gaussian Splatting (VBGS) enables continual learning without replay buffers using Coordinate Ascent Variational Inference (CAVI), but its per-frame iterations over all observed points make it too slow for real-time use with strict memory and latency requirements. We present ImprovedVBGS, an accelerated framework for on-the-fly continual reconstruction. This is achieved primarily through (i) spatially truncated variational inference, and (ii) improved reassignment that uses forwarding, truncation, and eliminates wasteful dynamic recompilation. On the NeRF synthetic dataset, we reduce mean per-frame latency from ~84.0\,s to ~0.050\,s on an RTX 3070 Ti, a 1680$\times$ speed-up. We also improve novel-view synthesis quality using an exact renderer with no added training costs.
\end{abstract}
\section{Introduction}
\paragraph{3D Gaussian Splatting.}
3D Gaussian Splatting (3DGS)~\cite{kerbl20233d} has become the standard for fast, high-quality novel-view synthesis~\cite{chen2024survey}. It represents a scene as a collection of 3D Gaussians, each parameterised by a mean position $\mu \in \mathbb{R}^3$, a covariance matrix $\Sigma$ (factorized into a scaling vector and rotation quaternion), an opacity $\alpha$, and a set of spherical harmonic coefficients for view-dependent colour. Given a set of posed training images, these parameters are optimized by backpropagating the gradient of a photometric loss (a weighted combination of $\mathcal{L}_1$ and D-SSIM) through differentiable tile-based rasterisation. 
\paragraph{Continual Learning for Gaussian Splatting.}
The 3DGS optimisation procedure assumes all training views are available simultaneously. With gradient-based approaches in a continual learning setting, where data arrives sequentially, means each new batch of images overwrites previously learned parameters, causing catastrophic forgetting. A common mitigation strategy, maintaining a replay buffer of past frames, increases both memory and compute costs with the number of observed views, making it poorly suited for resource-constrained scenarios. Continual on-device mapping, however, requires models that update incrementally without catastrophic forgetting~\cite{french1999catastrophic} or expensive replay buffers~\cite{matsuki2024gaussian,keetha2024splatam}.
Gradient-based methods for continual 3DGS typically rely on replay or local updates. Examples include local optimization in CL-Splats~\cite{ackermann2025clsplats} and generative replay in GaussianUpdate~\cite{zeng2025gaussianupdate}. These approaches either store past observations (replay) or perform additional optimization on changed regions, increasing compute and/or storage. Moreover, they mitigate catastrophic forgetting by revisiting or constraining past data rather than eliminating it by design.

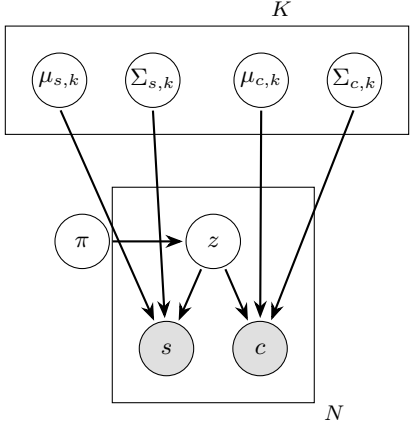
\begin{figure}[h]
\centering
\begin{tikzpicture}[
  >=Stealth,
  font=\small,
  latent/.style={circle, draw, minimum size=7.2mm, inner sep=0.5pt},
  obs/.style={circle, draw, fill=black!12, minimum size=7.2mm, inner sep=0.5pt},
  arr/.style={->, thick, shorten >=1pt, shorten <=1pt},
]
  \node[latent] (mus) {$\mu_{s,k}$};
  \node[latent, right=5mm of mus] (sigs) {$\Sigma_{s,k}$};
  \node[latent, right=7mm of sigs] (muc) {$\mu_{c,k}$};
  \node[latent, right=5mm of muc] (sigc) {$\Sigma_{c,k}$};
  \node[draw, fit=(mus)(sigs)(muc)(sigc), inner sep=3.5mm,
        label={[font=\footnotesize]above right:$K$}] (compplate) {};

  \node[latent, below=14mm of sigs, xshift=8mm] (z) {$z$};
  \node[obs, below left=9mm and 1mm of z] (s) {$s$};
  \node[obs, below right=9mm and 1mm of z] (c) {$c$};
  \node[draw, fit=(z)(s)(c), inner sep=3.5mm,
        label={[font=\footnotesize]below right:$N$}] (dataplate) {};

  \node[latent, left=10mm of z] (pi) {$\pi$};

  \draw[arr] (pi) -- (z);
  \draw[arr] (z) -- (s);
  \draw[arr] (z) -- (c);
  \draw[arr] (mus.south) -- (s);
  \draw[arr] (sigs.south) -- (s);
  \draw[arr] (muc.south) -- (c);
  \draw[arr] (sigc.south) -- (c);
\end{tikzpicture}
\caption{\textbf{VBGS generative model.}
Each of $N$ points comprises observed position $s$ and colour $c$
(shaded), generated by a latent component assignment $z\sim\mathrm{Cat}(\pi)$.
Component $k$ contributes multivariate Gaussian parameters
$(\mu_{s,k},\Sigma_{s,k})$ and $(\mu_{c,k},\Sigma_{c,k})$
(white $=$ unobserved). Adapted from~\cite{vbgs2024}.}
\label{fig:vbgs-gen}
\end{figure}

\paragraph{Variational Bayes Gaussian Splatting.}
Instead of using backpropagation and photometric error to train a scene model, VBGS~\cite{vbgs2024} formulates the problem as variational inference over a probabilistic mixture model. It models the scene as a generative mixture model where spatial position and colour are drawn from 3D Gaussians conditioned on latent component assignments (Fig.~\ref{fig:vbgs-gen}). The method uses conjugate priors (Normal-Inverse-Wishart on position $s$ and colour $c$ and Dirichlet on mixture weights $\pi$) and derives a closed-form variational update rule. Crucially, these updates are order-invariant and accumulate new observations by adding sufficient statistics to the posterior parameters, which makes VBGS inherently immune to catastrophic forgetting. In contrast to the methods above, VBGS requires neither a replay buffer to revisit past observations nor iterative local optimization on changed regions. 

Unlike traditional 3DGS, VBGS requires depth input for training. It also uses over $2\times$ more parameters (29 vs. 14), requiring more memory to model the same scene, and does not model view-dependent colour through spherical harmonics. The significant memory usage makes the original implementation infeasible for edge devices and consumer graphics cards. Another significant limitation, and the one this paper focuses on, is training time. The per-update assignment step requires evaluating the responsibility of each of the components for all $n$ observed data points ($O(nK)$), making the cost grow linearly with scene size. 

The follow-up work~\cite{zaino2026edge} makes multiple improvements, enabling edge applications by reducing training times and memory usage from ~234 minutes and 9.44 GB to ~61 minutes and 1.1 GB on an A5000 (24GB), and achieving novel view synthesis (NVS) training on a Jetson Orin Nano (8 GB). This is enabled by two key optimizations: kernel fusion for weighted sufficient statistics and automatic mixed-precision search. However, Zaino et al. still process all observed points per update, leaving per-frame latency far too high for true on-the-fly reconstruction.

\paragraph{ImprovedVBGS.} This work builds on the two prior studies by accelerating VBGS training for on-the-fly reconstruction. We reproduce the fused-statistics and mixed-precision from ~\cite{zaino2026edge} as an implementation baseline. We additionally add:
\begin{itemize}
  \item \textbf{Spatially truncated variational E-step:} KD-tree nearest-
        neighbour candidate pruning.
  \item \textbf{Truncated-ELBO reassignment:} reuse low-ELBO candidates from
        the truncated E-step instead of a dense ELBO rescan, with static-shape
        padding that preserves the original fraction-of-unused policy while
        eliminating dynamic-shape JAX recompilation.
\end{itemize}

\section{Method}
\label{sec:method}

\subsection{Fused sufficient-statistics contraction and mixed-precision search}
Following~\cite{zaino2026edge} we fuse memory-dominant kernels to eliminate large intermediate tensors and implement automatic mixed-precision search assigning operation-level precision under explicit numerical-stability constraints. However, as noted later, mixed-precision does not improve the performance once the spatially truncated E-step is applied.




\subsection{Spatially truncated variational E-step}

We want to find a distribution over the model's parameters ($\boldsymbol{\mu}, \boldsymbol{\Sigma}, \boldsymbol{c}, \boldsymbol{\pi}$) that best explains the observed data. Since computing the marginal log likelihood directly is intractable, we estimate the posterior distribution by maximising the Evidence Lower Bound (ELBO).
\begin{multline}
\text{ELBO} =
  \sum_{n=1}^N \big(
    \mathbb{E}_{q}[\log p(s_n | z_n, \mu_s, \Sigma_s)] \\
    + \mathbb{E}_{q}[\log p(c_n | z_n, \mu_c, \Sigma_c)]
    + \mathbb{E}_{q}[\log p(z_n | \pi)]
  \big)
\end{multline}
This is done using Coordinate Ascent Variational Inference (CAVI)~\cite{bishop2006pattern}~\cite{blei2017variational}. The first step
of this, the Variational E-step, computes the expectation over component
assignments for each data point $x_n$.
\begin{multline}
\log \gamma_{n,k} \propto
  \underbrace{\mathbb{E}_{q(\mu_{k,s},\Sigma_{k,s})}[\log p(s_n | \mu_{k,s}, \Sigma_{k,s})]}_{\text{spatial likelihood}} \\
  + \underbrace{\mathbb{E}_{q(\mu_{k,c},\Sigma_{k,c})}[\log p(c_n | \mu_{k,c}, \Sigma_{k,c})]}_{\text{color likelihood}}
  + \underbrace{\mathbb{E}_{q(\pi)}[\log \pi_k]}_{\text{mixture weight}}
\end{multline}

For models with a large number of components, computing this for each data point becomes a bottleneck. Spatial likelihood concentrates mass on nearby means; we show that a spatially truncated E-step (Alg.~\ref{alg:truncated}) preserves reconstruction quality while significantly improving training speed. Once per frame the pipeline builds a KD-tree $T$ on spatial means $\{\mu_{s,k}\}$ and queries $C$ nearest-neighbour component indices per point, stored as matrix $\mathcal{C}$. It keeps the closest means by Euclidean distance and evaluates the log-scores $\log\hat{\gamma}_{n,k}$ on that subset only, then forms soft responsibilities $R_n=\mathrm{softmax}(\log\hat{\gamma}_n)$ and $\mathrm{ELBO}_n=\mathrm{logsumexp}(\log\hat{\gamma}_n)$.

\begin{algorithm}[H]
\caption{Truncated \algfn{compute\_elbo\_delta} (spatially truncated E-step)}
\label{alg:truncated}
\begin{algorithmic}[1]
\small
\Require Batch $X_B=\{x_n\}$, model $\mathcal{M}_0$, candidates $\mathcal{C}_B$
\For{each point $x_n\in X_B$}
  \State $I_n \leftarrow {C}_B[n] $
  \State $\log\hat{\gamma}_n \leftarrow \mathrm{ELL}_{\mathrm{space}}+\mathrm{ELL}_{\mathrm{color}}+\mathbb{E}_{q(\pi)}[\log\pi]_{I_n}$
  \State $R_n \leftarrow \mathrm{softmax}(\log\hat{\gamma}_n)$
  \State $\mathrm{ELBO}_n \leftarrow \mathrm{logsumexp}(\log\hat{\gamma}_n)$
\EndFor
\State \Return $(\mathrm{ELBO},\,R)$
\end{algorithmic}
\end{algorithm}

\subsection{Improved reassignment}

\paragraph{Explanation of reassignment.}
Reassignment is necessary in the continual learning setting. The reassignment step relocates unused components to poorly modelled regions. The authors of VBGS~\cite{vbgs2024} demonstrate the importance of the reassignment with random initialisation, when the data statistics are not known ahead of time. In this step of the algorithm unused components are identified by checking the Dirichlet posterior parameters of the mixture weights. 

Each component $k$ has weight parameter $\alpha_k$. If a component is never assigned any points during the variational E-step this value decays down to the prior floor. At each step we use 5\% of the components taking the lowest $\alpha_k$ values. These are moved to poorly modelled regions identified using the per-point ELBO. 

Taking every point $x_n$ in the current batch, we compute its ELBO over the components of the mixture model. Let $ \log {\hat{\gamma}_{n,k}}$ be the unnormalised log-assignment score of point $n$ under component $k$. The per-point ELBO is given by the log-normaliser of the responsibilities:
\begin{equation}
\text{ELBO}_n = \log{ \sum_{k=1}^K \exp(\log \hat{\gamma}_{n,k})}
\end{equation}
A low $\text{ELBO}_n$ means the point has low probability under all current components. We select the $n_{\mathrm{reassign}}$ points with the lowest scores as reassignment destinations.

\paragraph{Truncated Reassignment and Reassignment Forwarding.} Conveniently, ELBO values are already computed in the E-step. The improved efficiency with truncation can also be applied here; we also find forwarding these values directly from the fit step by re-ordering the steps to be effective even under truncation. In the default continual order, reassignment runs before the variational EM step so relocated components receive same-frame evidence. Forwarding trades a small PSNR drop for lower latency. (Table~\ref{tab:lego})
\paragraph{Static Tensor Padding.} Improvements are also made by avoiding costly JAX recompilation caused by dynamic tensor shapes. The value $n_{\mathrm{reassign}}$ changes every frame, forcing recompilation. To prevent this we fix the reassignment array size across frames. Because $n_{\mathrm{reassign}}$ varies dynamically per frame, naive execution forces JAX to recompile execution graphs on each iteration. We eliminate this overhead by padding reassignment indices to a static compile-time shape $n_{\max}=\lfloor f\cdot N \rfloor$.

\subsection{Exact Rendering}

As the training processes does not use gradients from photometric loss the models trained using VBGS are not optimised for the quirks of any Gaussian rendering technique. Standard 3DGS rendering relies on approximating 3D Gaussians as a 2D projection which degrades accuracy. This method can benefit from an exact rendering technique such as 3DGEER \cite{3dgeer}, but unlike gradient-based methods changing the renderer does not impact training times.

\section{Results}
\label{sec:experiments}

\subsection{Setup}

All measurements were taken on a single desktop using an RTX~3070~Ti (8\,GB VRAM). This is substantially more constrained than the NVIDIA A5000 (24 \, GB) used in~\cite{zaino2026edge}. We implement the techniques that allowed them to make significant improvements ($234{\rightarrow}61$\,min, $9.44{\rightarrow}1.11$\,GB) but do not claim identical reproduction without the authors' unreleased implementation or the hardware to replicate their values. 

\subsection{NeRF Synthetic Dataset}
We cover all eight scenes of the \textbf{NeRF Synthetic dataset} ~\cite{mildenhall2020nerf} using the  200 training frames, 100 validation frames, using $N{=}10^5$ components with random initialisation. The reported PSNR, averaged across all validation frames is inline with the results reported by \cite{vbgs2024} on all scenes.
\begin{table}[h]
  \centering
  \caption{\textbf{NeRF Synthetic} (RTX~3070~Ti, random init, static-shape reassign before fit). Latency is train-loop seconds per frame. PSNR values are reported as $ \mu \pm \sigma$ over validation.}
  \label{tab:all-scenes}
  \footnotesize
  \setlength{\tabcolsep}{4pt}
  \begin{tabular}{@{}l c cc@{}}
  \toprule
  & & \multicolumn{2}{c}{\textbf{PSNR (dB)}} \\
  \cmidrule(l){3-4}
  \textbf{Scene} & \textbf{Latency (s/frame)} & \textbf{3DGS} & \textbf{3DGEER} \\
  \midrule
  chair     & 0.128 & $22.11\,{\pm}\,0.91$ & $22.50\,{\pm}\,0.98$ \\
  drums     & 0.132 & $18.65\,{\pm}\,0.41$ & $18.89\,{\pm}\,0.41$ \\
  ficus     & 0.117 & $21.26\,{\pm}\,0.68$ & $21.52\,{\pm}\,0.69$ \\
  hotdog    & 0.143 & $23.66\,{\pm}\,0.84$ & $23.95\,{\pm}\,0.83$ \\
  lego      & 0.136 & $21.89\,{\pm}\,0.77$ & $22.17\,{\pm}\,0.77$ \\
  materials & 0.133 & $20.66\,{\pm}\,1.45$ & $20.74\,{\pm}\,1.46$ \\
  mic       & 0.117 & $24.04\,{\pm}\,0.60$ & $24.33\,{\pm}\,0.63$ \\
  ship      & 0.159 & $21.52\,{\pm}\,0.81$ & $21.64\,{\pm}\,0.82$ \\
  \midrule
  \textbf{Mean} & \textbf{0.133} & $\mathbf{21.72\,{\pm}\,0.81}$ & $\mathbf{21.97\,{\pm}\,0.82}$ \\
  \bottomrule
  \end{tabular}
\end{table}
  
\subsection{Ablation Study}
The cumulative ablation study in Table ~\ref{tab:lego} validates each component of ImprovedVBGS. The combination of mixed precision and fused stats reached 25.6\,s/frame without reassignment but does not yield constructive gains when combined with truncation so is excluded from the table. Reassign forwarding provides a small speed up at the expense of some accuracy. This is because after truncation the \algfn{compute\_elbo\_delta} no longer dominates the per-frame latency as it does in the baseline, however the knowledge that the majority of the quality gains 21.09 to 22.04 can be obtained with only one \algfn{compute\_elbo\_delta} per training iteration may prove useful for future work. With $N{=}10^5$ components a batch size of 100 is initially chosen to avoid an OOM error, after optimisations are applied a batch size of 250,000 is used to minimise latency. 

\begin{table}[H]
\centering
\caption{\textbf{NeRF Lego Ablation.}}
\label{tab:lego}
\footnotesize
\setlength{\tabcolsep}{2pt}
\resizebox{\columnwidth}{!}{%
\begin{tabular}{@{}lcccc@{}}
\toprule
\textbf{Configuration} & \textbf{Batch Size} & \textbf{Latency (s/frame)} & \textbf{PSNR (dB)} \\
\midrule
Baseline VBGS ~\cite{vbgs2024} & 100 & 84.0 & $21.09\,{\pm}\,1.01$ \\
+ Fused Stats (~\cite{zaino2026edge}) & 100 & 41.0 & $21.09\,{\pm}\,1.01$ \\
+ Truncated E-step & 100 & 3.39 & $21.09\,{\pm}\,1.01$ \\
+ Large Batch Size & 250k & \textbf{0.050} & $21.09\,{\pm}\,1.01$ \\
+ Exact Rendering & 250k & 0.050 & $21.20\,{\pm}\,1.06$ \\
\midrule
+ Reassignment & 250k & 18.1 & $22.15\,{\pm}\,0.77$ \\
+ Truncated Reassignment & 250k & 0.373 & $22.15\,{\pm}\,0.77$ \\
+ Static Tensor Padding & 250k & 0.131 & $22.15\,{\pm}\,0.76$ \\
+ Reassign Forwarding & 250k & \textbf{0.107} & $22.04\,{\pm}\,0.80$ \\
\bottomrule
\end{tabular}%
}
\end{table}

\subsection{Latency analysis}
\label{sec:latency}

\begin{figure}[h]
  \centering
  \includegraphics[width=\linewidth]{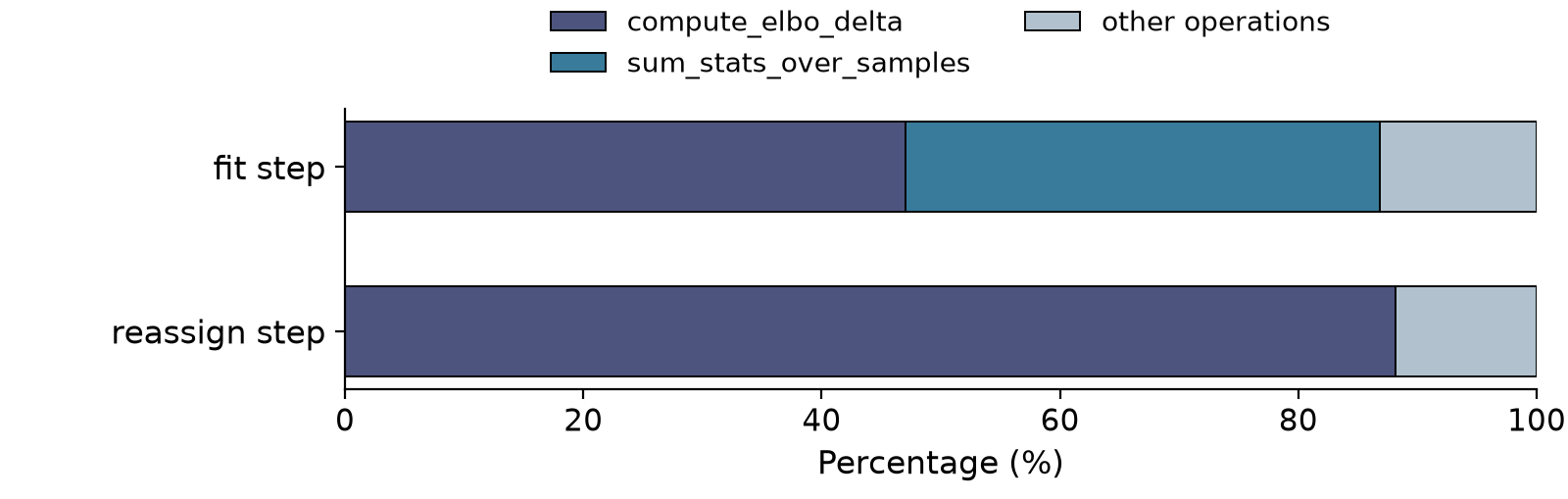}\\[0.35em]
  {\footnotesize (a) Baseline VBGS ~\cite{vbgs2024} + Fused stats ([11])}\\[0.6em]
  \includegraphics[width=\linewidth]{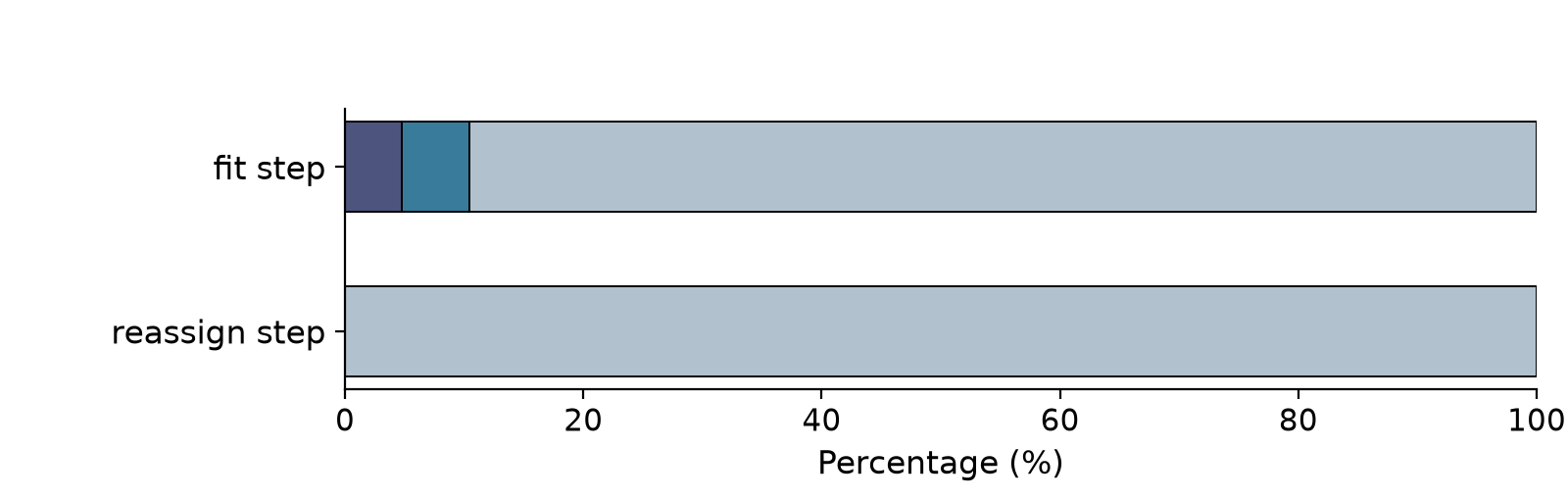}\\[0.35em]
  {\footnotesize (b) Ours }
  \caption{ \textbf{Latency analysis.} Fit and reassignment steps with latency composition on Blender Lego
  ( $N{=}10^5$, fp64, fused stats)}
  \label{fig:fit-reassign}
\end{figure}

Figure~\ref{fig:fit-reassign}  shows the composition of the reassign and fit steps on
Lego ($N{=}10^5$, fp64; dense batch~$100$, truncated $C{=}4$
batch~$250$k).  The baseline fit step is dominated by \algfn{compute\_elbo\_delta} $28.8$\,s/frame (47\%) and \algfn{sum\_stats\_over\_samples} $24.3$\,s/frame (40\%),
with the rest of the work taking only $8.1$\,s/frame. In ImprovedVBGS they contribute to a small part of the total time taking only  $3.1$\,ms/frame (4.8\%) and $3.7$\,ms/frame (5.6\%) each with other operations taking $58.4$\,ms/frame. 

For the reassignment step the baseline composition is: $22.7$\,s/frame (88\%) for \algfn{compute\_elbo\_delta}. In this step we are able to remove the \algfn{compute\_elbo\_delta} leaving only the other operations which take $107$\,ms/frame.

\section{Conclusion}
ImprovedVBGS is an open implementation of VBGS ~\cite{zaino2026edge}, ~\cite{vbgs2024} incorporating a  faster truncated E-step and lightweight reassignment. As summarised in Table~\ref{tab:lego} ImprovedVBGS keeps replay-free continual learning intact while making VBGS training practical and timely on consumer hardware.

{\small
\bibliographystyle{ieeenat_fullname}
\bibliography{refs}
}

\clearpage
\onecolumn
\appendix

\section{Additional Qualitative Results}
\noindent Figure~\ref{fig:all_scenes} provides additional visual comparisons between the ground truth and our ImprovedVBGS predictions across all eight scenes of the NeRF Synthetic dataset (static-shaped reassignment protocol).

\vspace{0.35em}
\noindent
\begin{center}
  \begin{minipage}{0.24\textwidth}
    \centering
    \includegraphics[width=\linewidth]{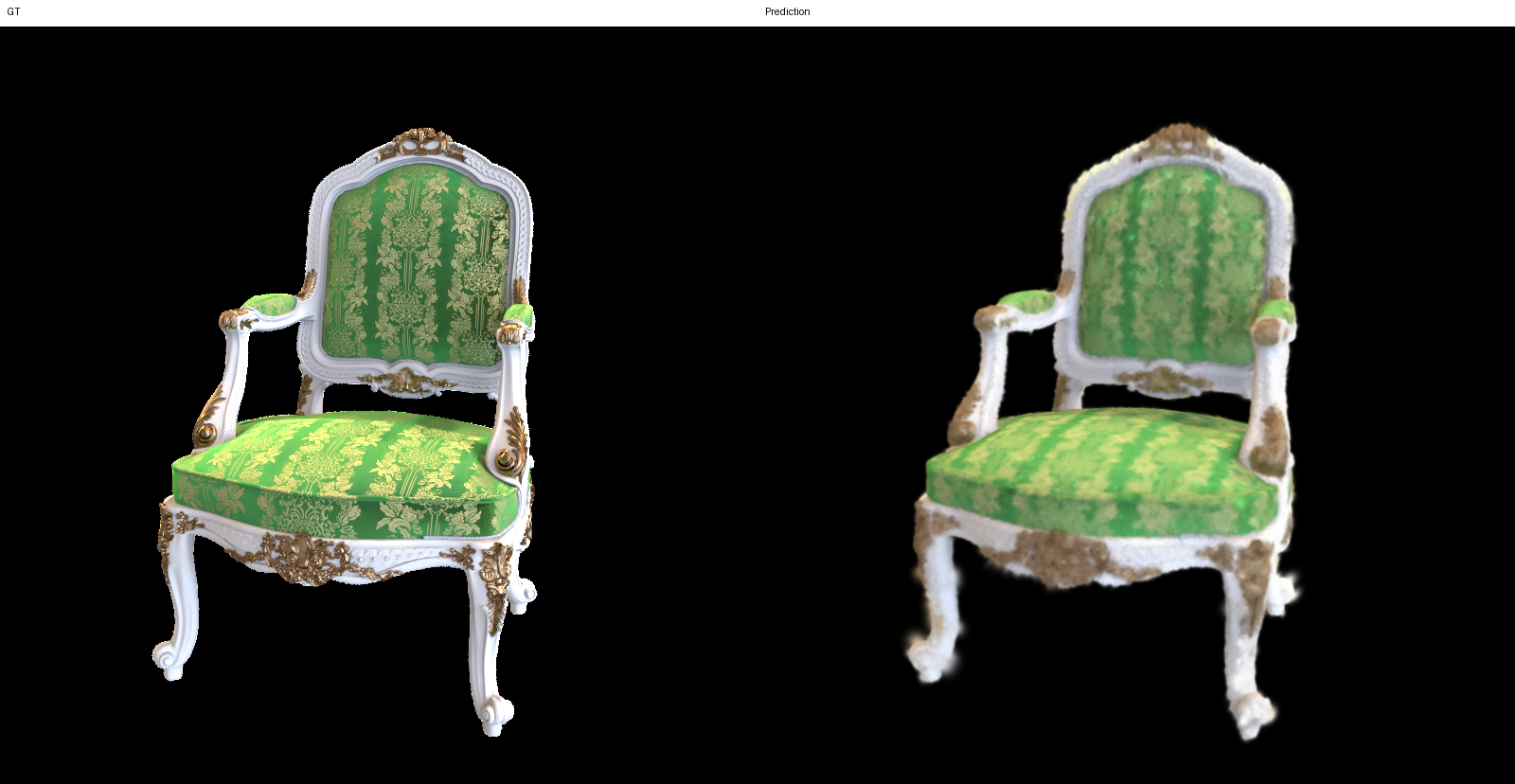}\\[-0.25em]
    {\footnotesize (a) Chair}
  \end{minipage}\hfill
  \begin{minipage}{0.24\textwidth}
    \centering
    \includegraphics[width=\linewidth]{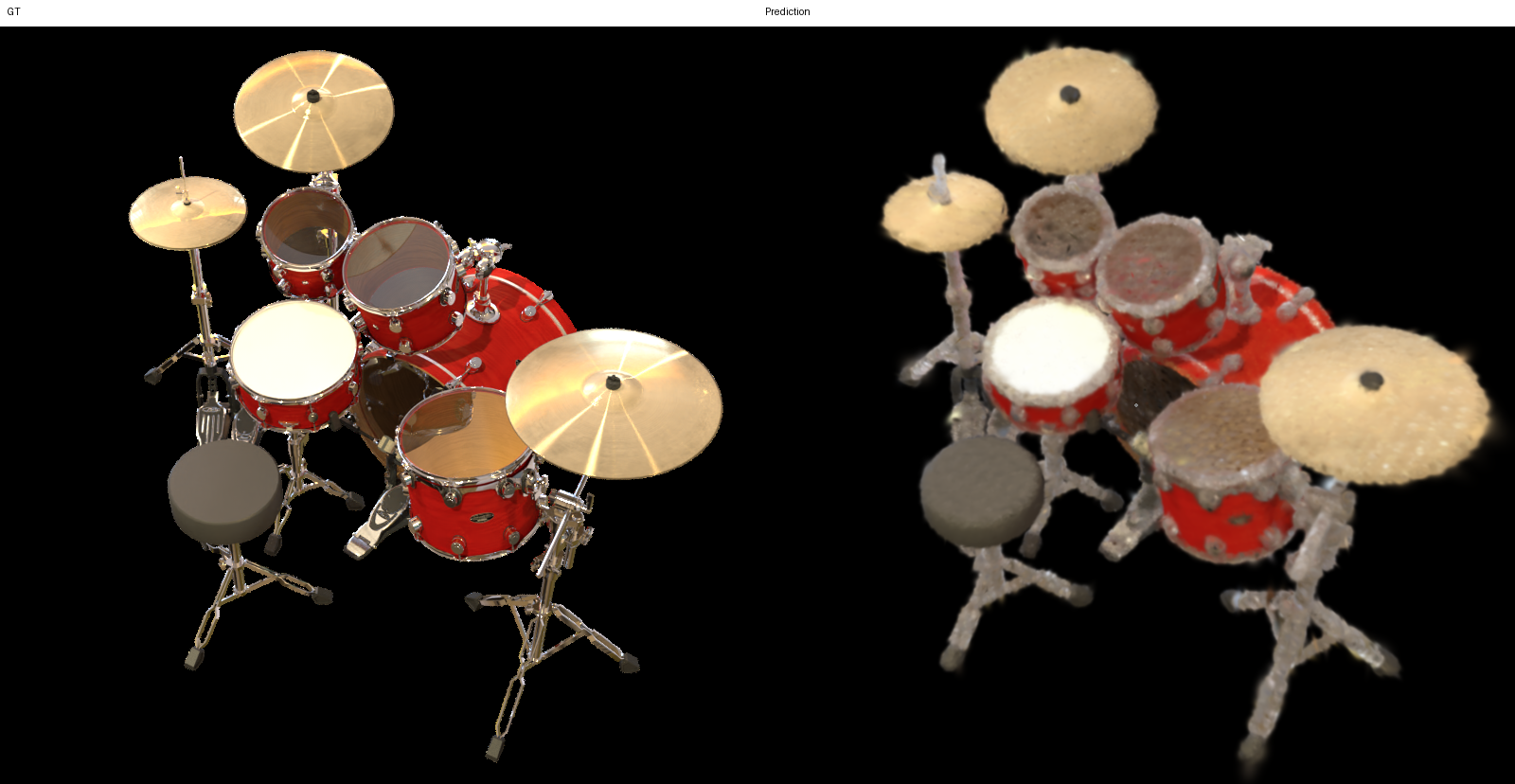}\\[-0.25em]
    {\footnotesize (b) Drums}
  \end{minipage}\hfill
  \begin{minipage}{0.24\textwidth}
    \centering
    \includegraphics[width=\linewidth]{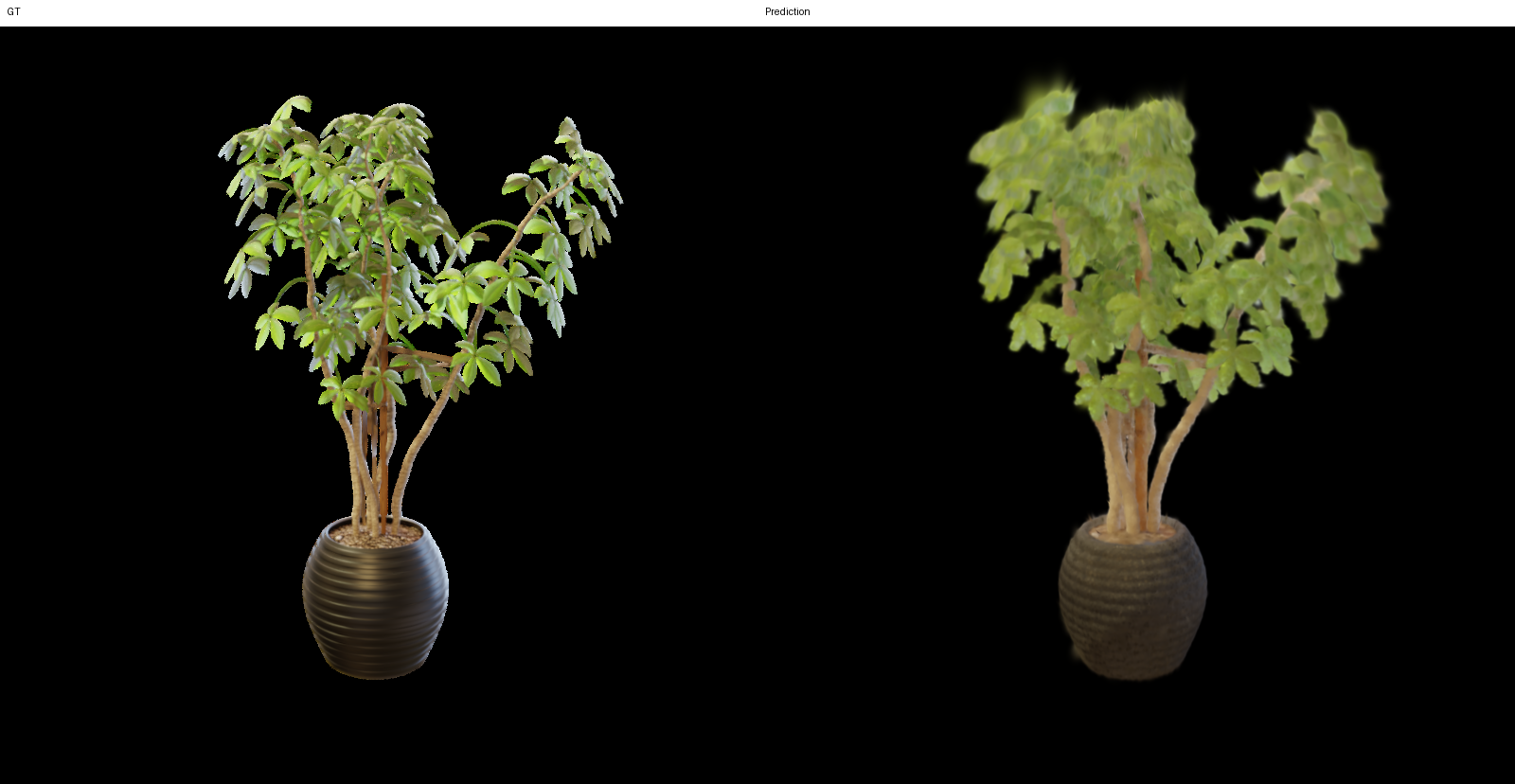}\\[-0.25em]
    {\footnotesize (c) Ficus}
  \end{minipage}\hfill
  \begin{minipage}{0.24\textwidth}
    \centering
    \includegraphics[width=\linewidth]{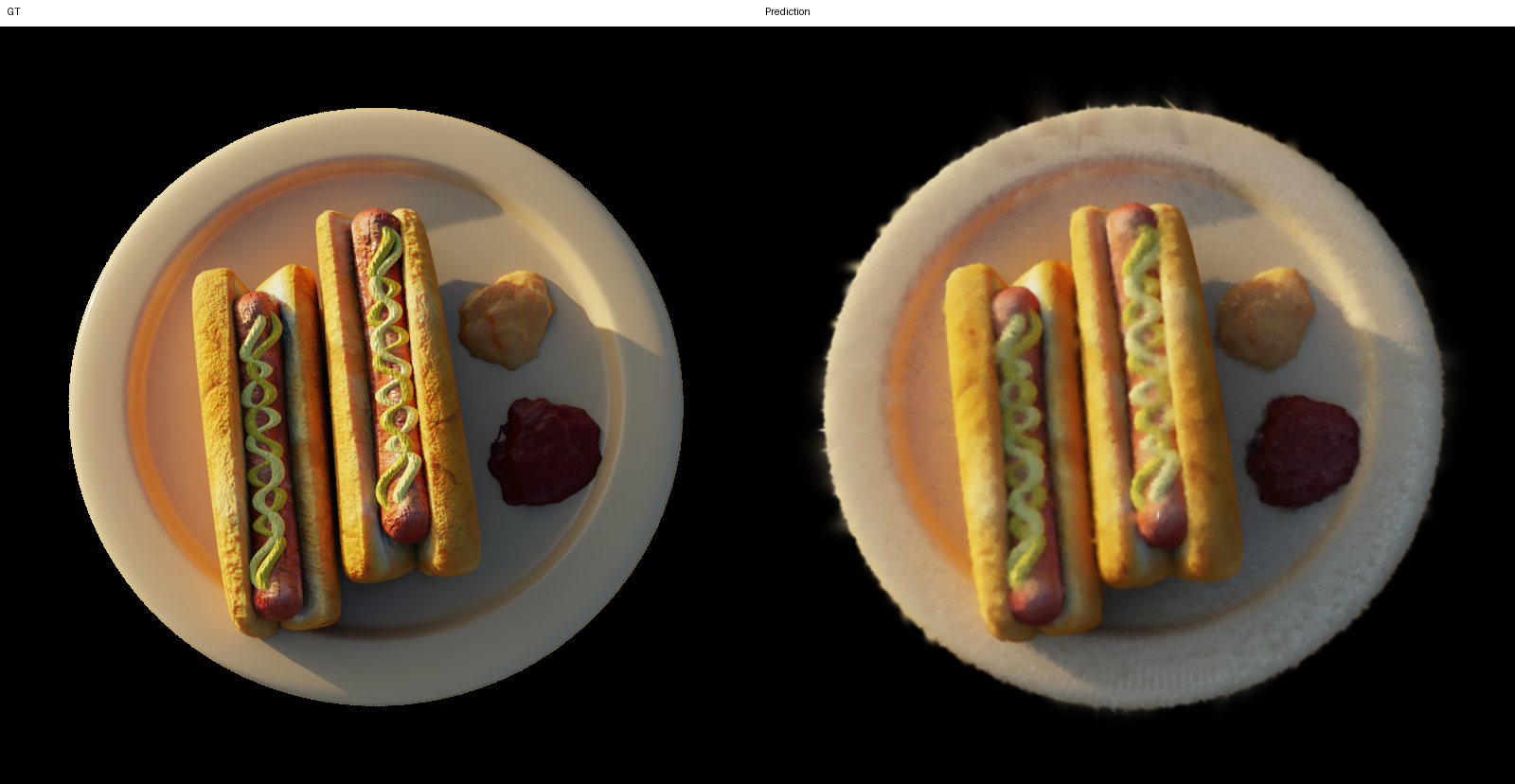}\\[-0.25em]
    {\footnotesize (d) Hotdog}
  \end{minipage}

  \vspace{0.25em}
  \begin{minipage}{0.24\textwidth}
    \centering
    \includegraphics[width=\linewidth]{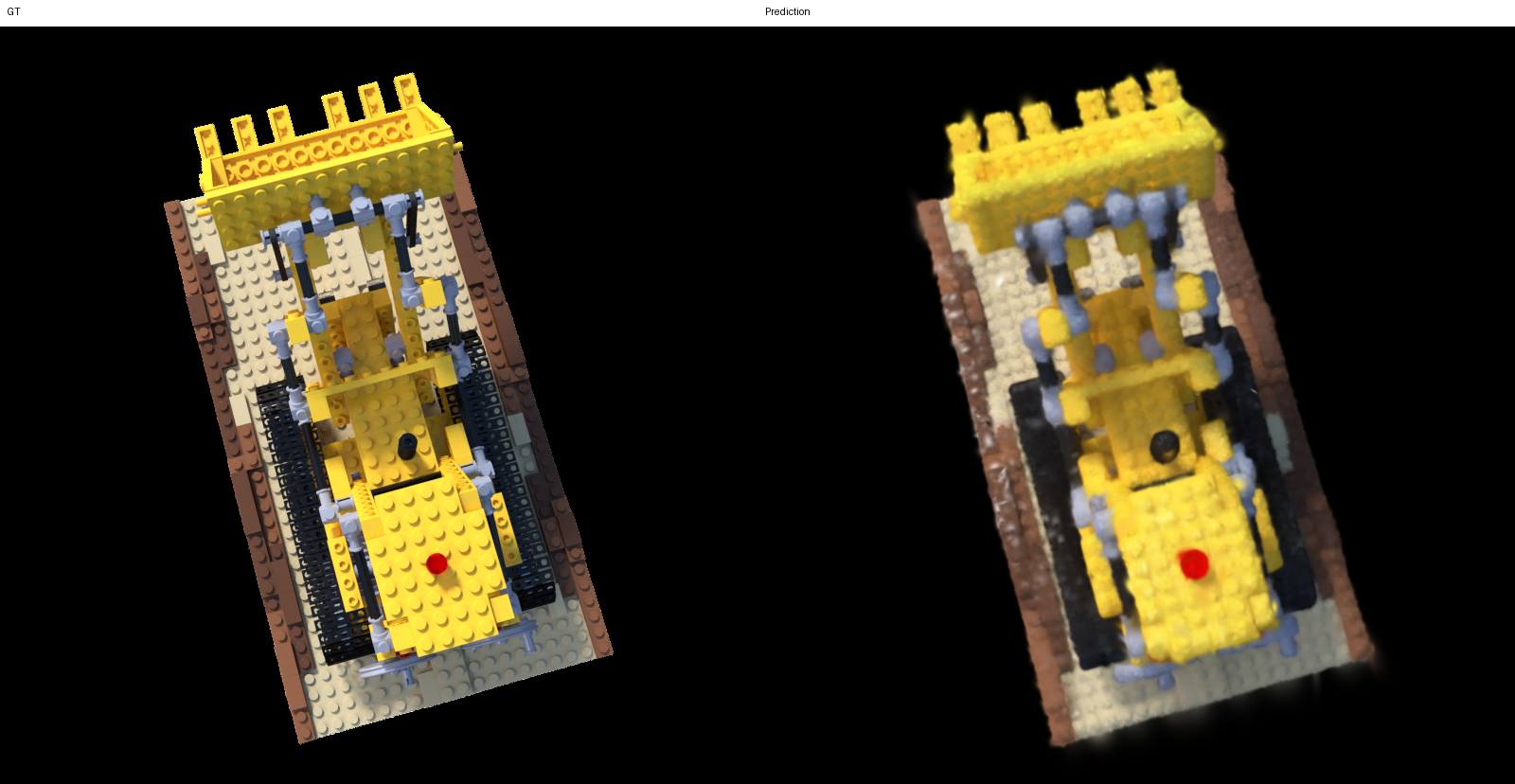}\\[-0.25em]
    {\footnotesize (e) Lego}
  \end{minipage}\hfill
  \begin{minipage}{0.24\textwidth}
    \centering
    \includegraphics[width=\linewidth]{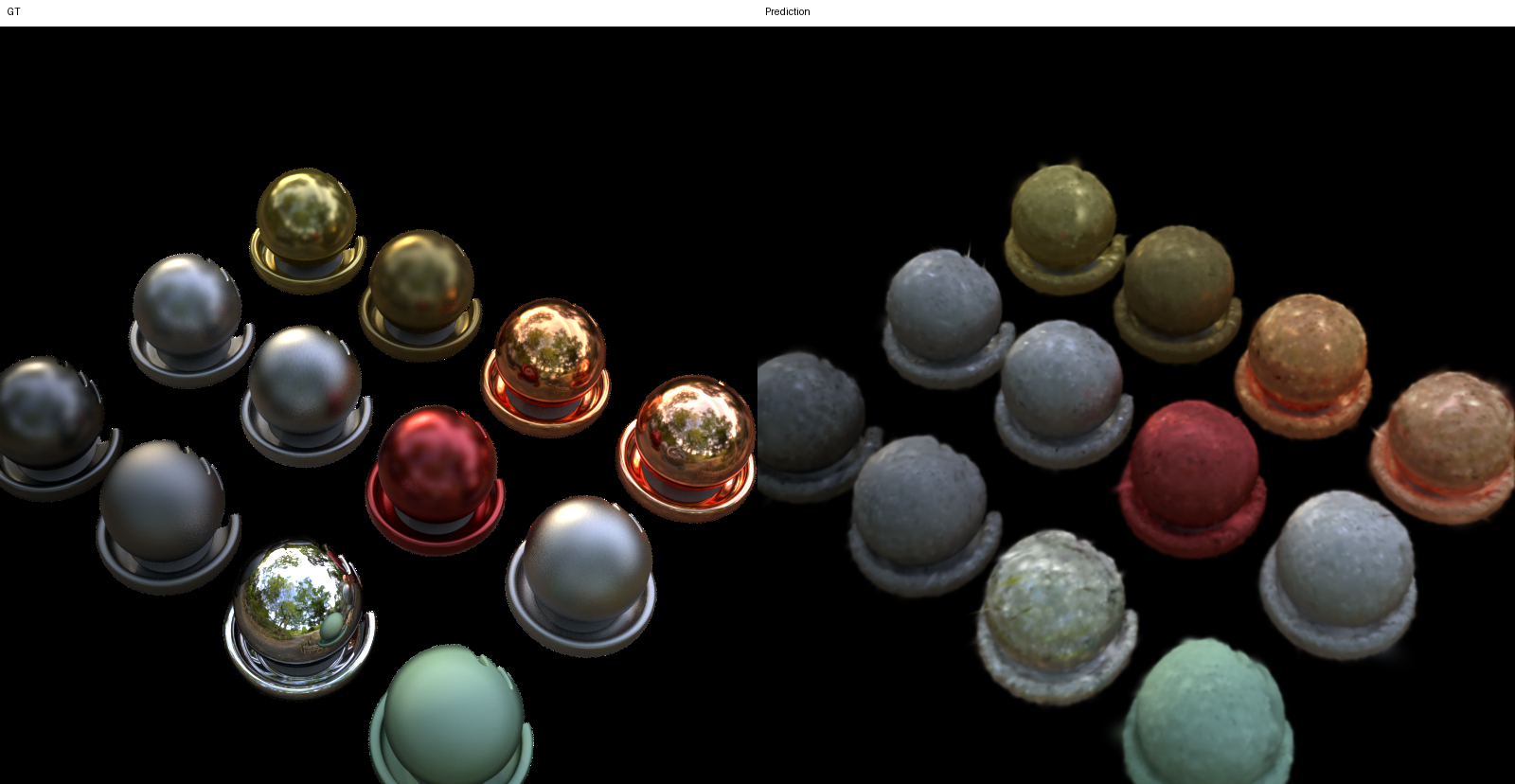}\\[-0.25em]
    {\footnotesize (f) Materials}
  \end{minipage}\hfill
  \begin{minipage}{0.24\textwidth}
    \centering
    \includegraphics[width=\linewidth]{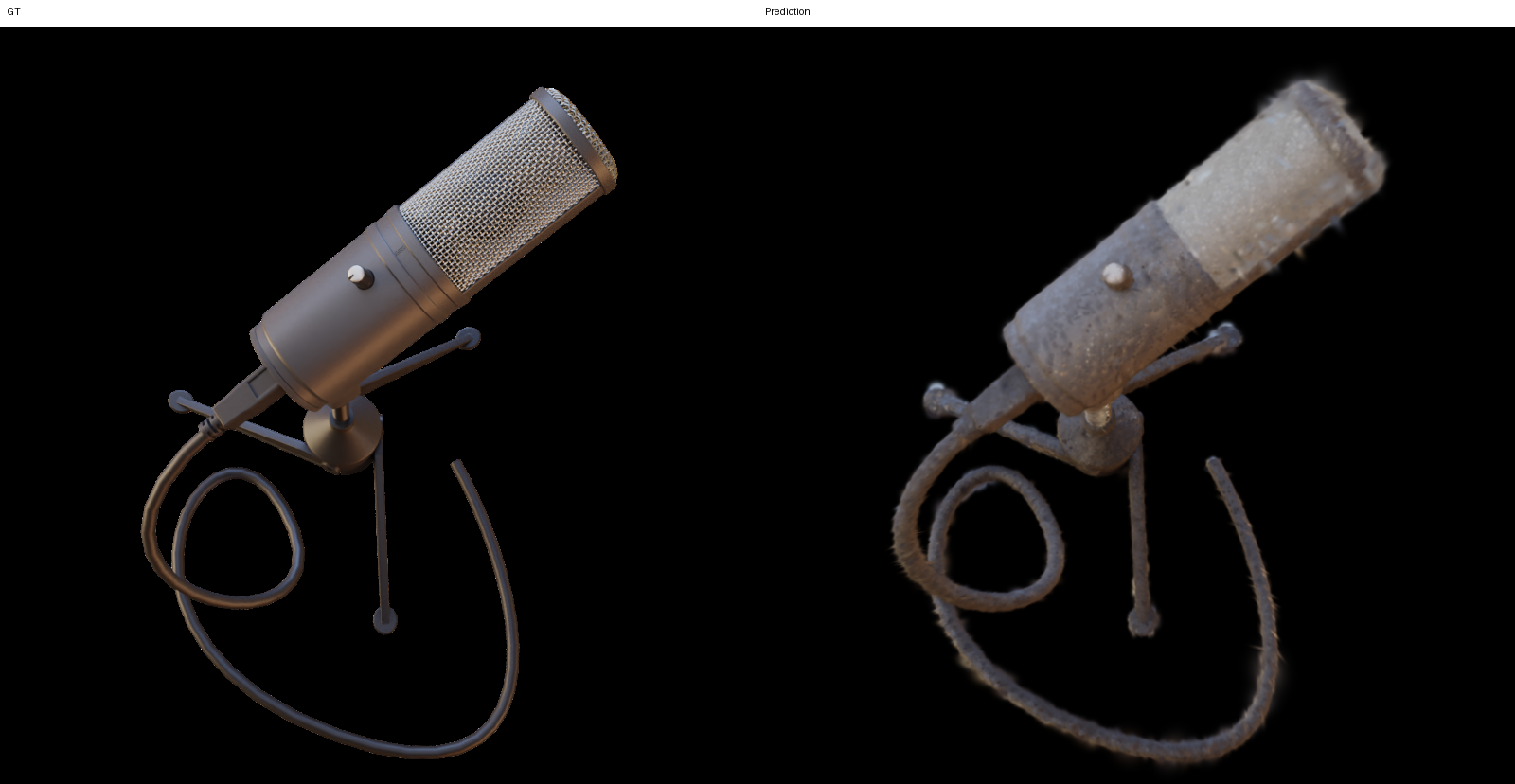}\\[-0.25em]
    {\footnotesize (g) Mic}
  \end{minipage}\hfill
  \begin{minipage}{0.24\textwidth}
    \centering
    \includegraphics[width=\linewidth]{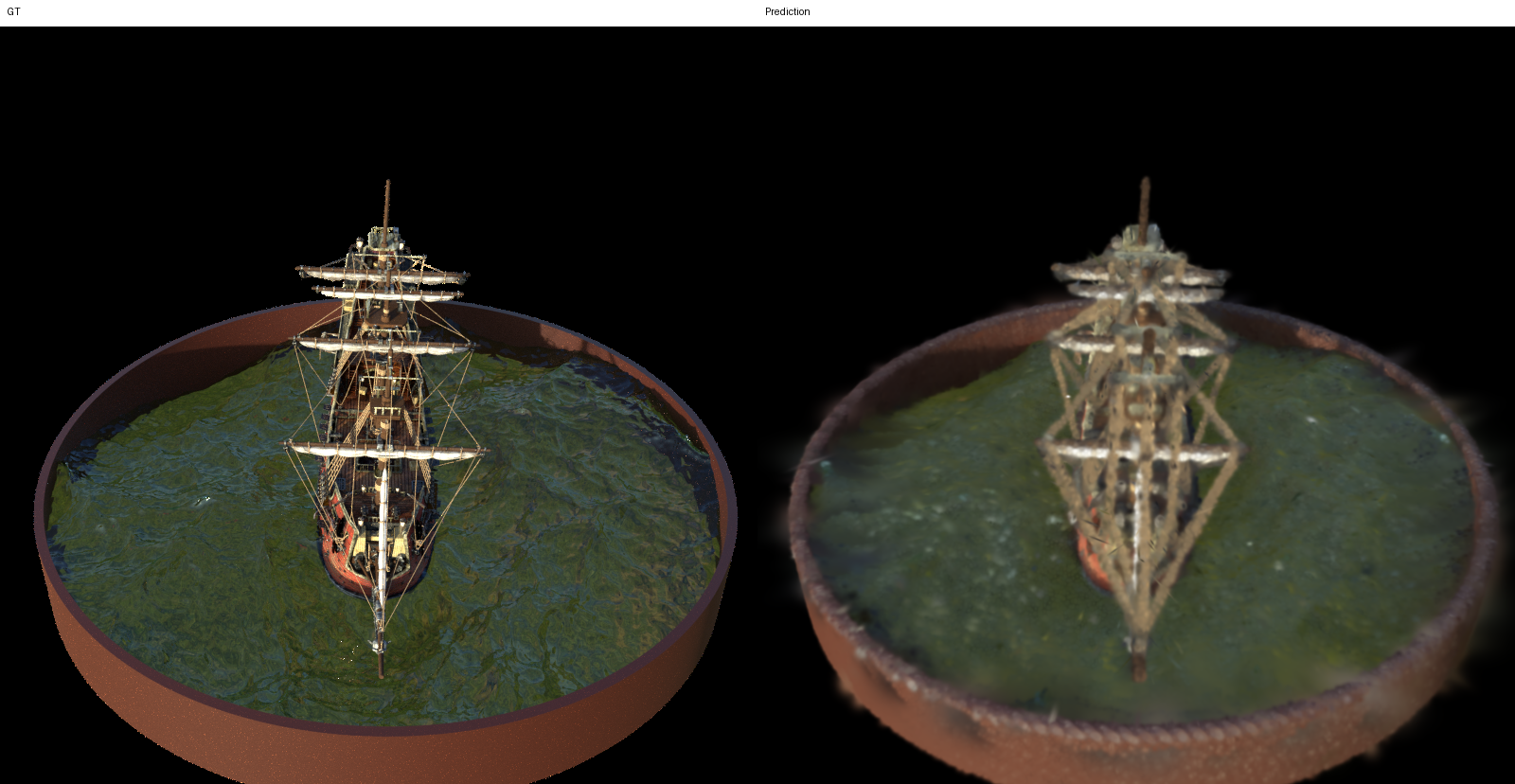}\\[-0.25em]
    {\footnotesize (h) Ship}
  \end{minipage}

  \vspace{0.35em}
  \captionof{figure}{Qualitative results on the NeRF Synthetic dataset with reassignment enabled. Each panel shows the Ground Truth (left) versus the ImprovedVBGS prediction (right) after 200 frames of continual training ($N{=}10^5$, Batch Size=250\,k,$C{=}4$).}
  \label{fig:all_scenes}
\end{center}

\begin{figure}[h]
  \centering
  \includegraphics[width=\columnwidth]{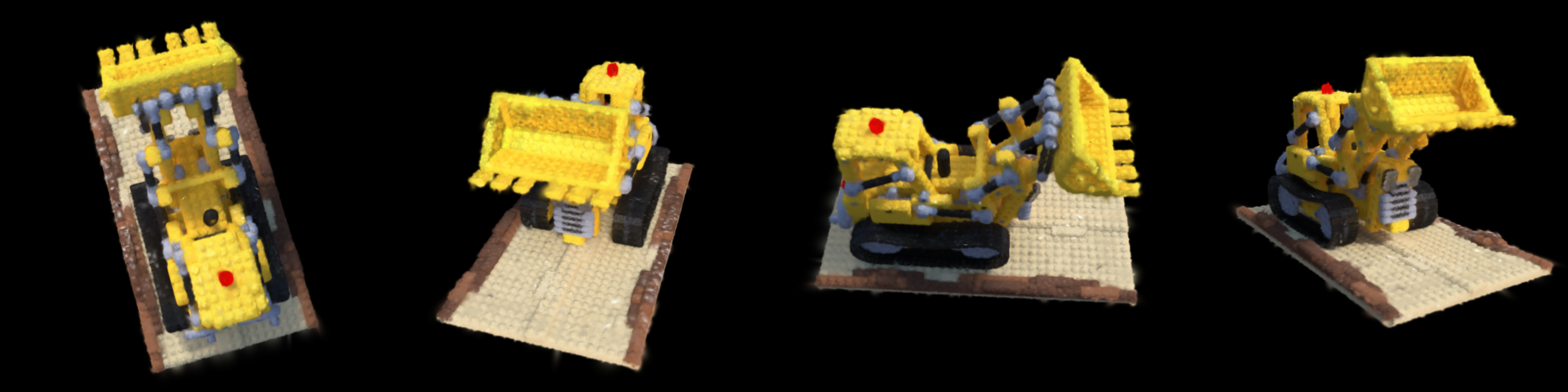}
  \caption{Lego validation predictions from the full system.}
  \label{fig:lego}
\end{figure}

\end{document}